\documentclass[runningheads]{llncs}

\usepackage{accv}

\usepackage{accvabbrv}

\usepackage[table]{xcolor}
\usepackage{graphicx} 
\usepackage{array} 
\usepackage{multirow} 
\usepackage{makecell} 
\usepackage{booktabs} 
\usepackage{arydshln} 
\usepackage{algorithm}
\usepackage{algpseudocode}
\usepackage{amsmath}
\usepackage{amssymb}
\usepackage{wrapfig} 
\usepackage{colortbl} 

\usepackage[accsupp]{axessibility}  

\usepackage[pagebackref,breaklinks,colorlinks,citecolor=accvblue]{hyperref}

\usepackage{orcidlink}

\begin{document}

\title{Not All Task Vectors Need Equal Rank: Energy-Proportional Allocation for Model Merging} 

\titlerunning{Not All Task Vectors Need Equal Rank}

\author{Hyunjoong Cho\inst{1} \and
Jinhyeok Jang\inst{2,3}\thanks{Corresponding author.}}

\authorrunning{H.~Cho and J.~Jang.}

\institute{LG CNS\and
ETRI \and
UST \\ \email{\{hyeunjoong, jangjh6297\}@gmail.com} }


\maketitle

\begin{abstract}
Model merging aims to combine multiple fine-tuned models derived from a common pretrained model into a single multi-task model without additional joint training. Recent spectral merging methods improve over simple weight averaging by exploiting low-rank structures of task-specific updates, but they commonly assign the same rank capacity to every task. This uniform allocation ignores that task vectors can have heterogeneous spectral complexity, causing the shared merging space to be used suboptimally. In this paper, we propose Spectral Energy-proportional Rank Allocation (SERA), a simple task-adaptive strategy that allocates ranks according to the singular-value energy structure of each task vector. By assigning richer spectral capacity to complex or isolated tasks and fewer directions to compact tasks, SERA extends SVD-based model merging from uniform-capacity merging to task-dependent capacity allocation. Experiments under standard vision model merging protocols show that SERA improves multi-task merging performance while preserving the same total rank budget as existing spectral merging methods. Further analysis demonstrates that task-level spectral concentration is closely related to the per-task effect of adaptive rank allocation, providing insight into when and why SERA is effective.
\end{abstract}

\section{Introduction}
\label{sec:intro}

Model merging aims to combine multiple fine-tuned models derived from a pretrained model into a single model. Instead of training a unified model from scratch on all downstream datasets, model merging reuses existing task-specialized models and aggregates their knowledge directly in parameter space. This makes model merging an attractive paradigm for building multi-task or more generalized models, especially in an era where a large number of pretrained and fine-tuned models are publicly available. By exploiting such models as reusable assets, merging provides a practical and efficient route toward model reuse without requiring access to the original training data or costly joint optimization.

The success of model merging relies on an implicit geometric assumption about the weight space around a pretrained model. Since all fine-tuned models are initialized from the same pretrained weights, their task-specific updates are expected to lie in a relatively smooth region of the loss landscape \cite{garipov2018loss}. Under this condition, interpolating or aggregating different task updates can preserve useful knowledge from each task while avoiding severe degradation. Another implicit assumption is that task knowledge is not entirely independent: the updates learned from different tasks are expected to be at least partially complementary or correlated. These assumptions explain why simple weight-space operations can often produce a merged model with non-trivial multi-task capability.

Early studies mainly explored direct weight averaging \cite{wortsman2022model}. For example, model soups showed that averaging independently fine-tuned models can improve robustness when the models share the same initialization and training configuration. Later methods moved beyond naive averaging by considering the geometry of model weights or task vectors. Distance-aware strategies \cite{jang2024model} attempted to account for the relative positions of fine-tuned models in weight space, while sign- or conflict-aware methods aimed to reduce destructive interference among task updates. More recent SVD-based approaches further introduced cross-parameter correlations by decomposing task vectors into structured low-rank components. These methods suggest that effective merging requires not only averaging parameter values but also preserving meaningful directions in the update space.

Among these directions, SVD-based merging is particularly relevant because it provides an explicit way to represent task updates through singular directions. By projecting task vectors onto low-rank subspaces, such methods can capture dominant update structures while suppressing noisy or redundant components. DC-Merge extends this idea by constructing a cover space from task-specific singular vectors and preserving directional consistency through a structured masking mechanism. This line of work shows that low-rank structure and cross-parameter relations are crucial for improving model merging beyond independent parameter-wise aggregation.

However, existing SVD-based merging methods still commonly treat all tasks as equally demanding. In particular, they typically allocate the same rank capacity to every task when constructing the shared merging space. This uniform treatment is simple, but it ignores the fact that different tasks may have very different structural properties in their task vectors. Some tasks can be represented compactly by a small number of dominant singular directions, while others require a broader set of directions to preserve their task-specific knowledge. In other words, task vectors may differ not only in magnitude or direction, but also in their intrinsic spectral complexity.

This observation motivates a different view of model merging: the rank capacity assigned to each task should reflect the structural complexity of that task. Difficult or isolated tasks may require richer low-rank representations in order to preserve their task-specific knowledge during merging. In contrast, tasks that are easier, more redundant, or more similar to other tasks may be sufficiently represented with a smaller rank. Uniform rank allocation can therefore be suboptimal because it may over-allocate capacity to compact tasks while under-representing complex or isolated ones.

In this paper, we propose Spectral Energy-proportional Rank Allocation (SERA), a task-adaptive strategy for spectral model merging. SERA estimates each task's rank demand from the singular-value structure of its task vector and reallocates a fixed cover-space rank budget accordingly. Instead of assigning the same rank to all tasks, SERA gives more spectral capacity to complex tasks and fewer directions to compact tasks, improving capacity usage without increasing the total rank budget.

Our method preserves the benefits of low-rank SVD-based merging while extending it from uniform-capacity merging to task-dependent capacity allocation. It exploits cross-parameter correlations through singular-vector representations and requires no additional training data or gradient-based optimization, relying only on the task vectors available at merging time.

Our contributions are summarized as follows:
\begin{enumerate}
    \item We reveal that uniform rank allocation in SVD-based model merging overlooks the heterogeneous spectral complexity of task-specific updates, leading to suboptimal use of the shared merging capacity.

    \item We propose Spectral Energy-proportional Rank Allocation, a task-adaptive merging strategy that allocates cover-space ranks according to the singular-value energy structure of each task vector while preserving the low-rank nature of existing SVD-based merging methods.

    \item We demonstrate that task-dependent rank allocation improves multi-task model merging and provide empirical analysis showing that the spectral structure of task vectors is closely related to the per-task behavior of the merged model.
\end{enumerate}






\section{Related Works}


\noindent \textbf{Weight-space model merging.}
Model merging aims to combine multiple models into a single model directly in weight space, without additional joint training. Early studies showed that simple weight averaging can be effective when the models share the same initialization and lie in a compatible loss basin. Model Soups \cite{wortsman2022model} demonstrated that averaging multiple fine-tuned models can improve accuracy and robustness without increasing inference. Fisher-weighted averaging further interpreted model merging from a Bayesian perspective and incorporated parameter importance through Fisher information~\cite{matena2022merging}. Model Stock extended this line of work by analyzing the geometry of fine-tuned weights and approximating a center-close solution using only a few fine-tuned models~\cite{jang2024model}. These methods establish weight-space averaging as a practical mechanism for model reuse, but they mainly focus on global weight aggregation and do not explicitly model the heterogeneous structure of task-specific updates.

\noindent \textbf{Task-vector-based merging.}
Task Arithmetic~\cite{ilharcoediting} introduced the task vector, defined as the difference between fine-tuned and pretrained weights, and showed that arithmetic operations on task vectors can steer model behavior and compose multiple task abilities. This formulation has become a common basis for multi-task model merging. However, directly adding task vectors often suffers from interference among tasks. TIES-Merging \cite{yadav2023ties} addressed this issue by removing small redundant updates, resolving sign conflicts, and merging only sign-consistent parameters~\cite{yadav2023ties}. AdaMerging instead learns task-wise or layer-wise merging coefficients using unlabeled data and entropy minimization, showing that different tasks and layers should not necessarily share the same merging coefficient~\cite{yang2024adamerging}. DARE \cite{yu2024dare} proposed a drop-and-rescale strategy for delta parameters to reduce redundancy before merging~\cite{yu2024dare}. Consensus Task Arithmetic (Consensus TA)~\cite{wang2024localizing} further localized task-specific information in sparse parameter supports and removed weights that are detrimental to multi-task fusion. 

\noindent \textbf{SVD-based model merging.}
Recent spectral merging methods have moved beyond parameter-wise aggregation by exploiting the low-rank structure of task updates~\cite{stoica2025model,marczak2025no,gargiulo2025task,zhang2026dc}. TSV shows that layer-wise task vectors contain meaningful singular directions and reduces interference by selecting and decorrelating task-relevant singular vectors~\cite{gargiulo2025task}. DC-Merge further constructs a shared cover space from task-wise singular directions and preserves directional consistency during aggregation~\cite{zhang2026dc}. These approaches demonstrate that singular-space representations can capture cross-parameter correlations that are overlooked by flat task-vector operations.

However, most spectral merging methods still use task-uniform or manually specified rank capacity, implicitly assuming that all task vectors require similar spectral budgets. This can be suboptimal when tasks have heterogeneous spectral complexity. Recent works have begun to address this limitation: AdaRank adaptively prunes singular components to reduce interference~\cite{lee2025adarank}, while PAVE purifies task vectors in a knowledge-aware subspace with spectral rank allocation~\cite{an2025purifying}. In contrast, our work focuses on a complementary problem: allocating a fixed cover-space rank budget across tasks before directional-consistent aggregation. SERA requires no additional training data or gradient-based optimization, and reallocates the same total rank budget solely according to the singular-value energy structure of the available task vectors.

\section{Problem Formulation}
\label{sec:problem_formulation}

Let $\{\theta_k\}_{k=1}^{K}$ be $K$ task-specific fine-tuned models initialized from the same pretrained model $\theta_{0}$. The task vector for task $k$ is defined as
\begin{equation}
    \Delta_k = \theta_k - \theta_{0} .
    \label{eq:task_vector}
\end{equation}
The goal is to construct a single merged model
\begin{equation}
    \theta^\star = \theta_{0} + \Delta^\star
    \label{eq:merged_model}
\end{equation}
that performs well across all $K$ tasks without access to the original training data. For each mergeable layer $l$, let $\Delta_k^{(l)}=\theta_k^{(l)} - \theta_0^{(l)}$ 
denote the layer-wise task delta. The same merging procedure is applied independently to each mergeable layer.

Existing merging methods, including weight averaging~\cite{wortsman2022model}, geometry-aware combinations~\cite{jang2024model}, and sign-aware task-vector aggregation~\cite{yadav2023ties}, construct $\Delta^\star$ through global or parameter-wise rules applied to the task vectors. SVD-based methods further exploit the spectral structure of each layer-wise task delta. For each task $k$ and layer $l$, we write the compact SVD as
\begin{equation}
    \Delta_k^{(l)}
    =
    U_{k,l} S_{k,l} V_{k,l}^{\top},
    \qquad
    S_{k,l}
    =
    \mathrm{diag}
    \left(
    s_{k,l,1}, \ldots, s_{k,l,q_l}
    \right),
    \qquad
    q_l = \min(m_l,n_l).
    \label{eq:compact_svd}
\end{equation}
Here, $U_{k,l} \in \mathbb{R}^{m_l \times q_l}$ and
$V_{k,l} \in \mathbb{R}^{n_l \times q_l}$ are the left and right singular-vector matrices. We denote their column vectors as
\begin{equation}
    U_{k,l}
    =
    [u_{k,l,1}, \ldots, u_{k,l,q_l}],
    \qquad
    V_{k,l}
    =
    [v_{k,l,1}, \ldots, v_{k,l,q_l}].
    \label{eq:singular_vector_columns}
\end{equation}
For any rank $r$, we define the top-$r$ singular-vector submatrices as
\begin{equation}
    U_{k,l}^{[1:r]}
    =
    [u_{k,l,1}, \ldots, u_{k,l,r}],
    \qquad
    V_{k,l}^{[1:r]}
    =
    [v_{k,l,1}, \ldots, v_{k,l,r}].
    \label{eq:top_r_singular_vectors}
\end{equation}

DC-Merge~\cite{zhang2026dc} constructs a shared cover space by concatenating the same number of top singular directions from each task. Specifically, with a uniform base rank $r_{\mathrm{base}}^{(l)}$, it forms
\begin{equation}
    \widetilde{U}_{\mathrm{DC}}^{(l)}
    =
    \left[
    U_{1,l}^{[1:r_{\mathrm{base}}^{(l)}]},
    \ldots,
    U_{K,l}^{[1:r_{\mathrm{base}}^{(l)}]}
    \right],
\qquad
    \widetilde{V}_{\mathrm{DC}}^{(l)}
    =
    \left[
    V_{1,l}^{[1:r_{\mathrm{base}}^{(l)}]},
    \ldots,
    V_{K,l}^{[1:r_{\mathrm{base}}^{(l)}]}
    \right].
    \label{eq:dcmerge_v_basis}
\end{equation}
Here, the brackets denote column-wise concatenation, and
$\mathrm{polar}(\cdot)$ denotes the orthonormal factor of the thin polar
decomposition. The cover matrices are obtained as
\begin{equation}
    C_U^{(l)}
    =
    \mathrm{polar}
    \left(
    \widetilde{U}_{\mathrm{DC}}^{(l)}
    \right),
    \qquad
    C_V^{(l)}
    =
    \mathrm{polar}
    \left(
    \widetilde{V}_{\mathrm{DC}}^{(l)}
    \right).
    \label{eq:dcmerge_cover_matrices}
\end{equation}

Each task delta is projected into the cover space, aggregated using a TIES-style sign-consensus operator~\cite{yadav2023ties}, and reconstructed. We denote this aggregation operator by $\mathcal{A}_{\mathrm{TIES}}$:
\begin{equation}
    M_k^{(l)}
    =
    C_U^{(l)\top}
    \Delta_k^{(l)}
    C_V^{(l)},
    \qquad
    \overline{M}^{(l)}
    =
    \mathcal{A}_{\mathrm{TIES}}
    \left(
    M_1^{(l)}, \ldots, M_K^{(l)}
    \right).
    \label{eq:cover_projection_aggregation}
\end{equation}
The merged update is reconstructed from the cover-space representation as
\begin{equation}
    \Delta^{\star(l)}
    =
    C_U^{(l)}
    \overline{M}^{(l)}
    C_V^{(l)\top}.
    \label{eq:cover_reconstruction}
\end{equation}

Although this cover-space formulation provides an effective way to aggregate task-specific updates, its merging capacity is assigned in a task-uniform manner. DC-Merge allocates the same number of singular directions $r_{\mathrm{base}}^{(l)}$ to every task when constructing the cover space. This uniform allocation ignores task-specific spectral complexity, namely how broadly each task update is distributed across singular directions. As a result, spectrally compact tasks may receive unnecessary rank capacity, whereas spectrally diffuse tasks may be under-represented under the same fixed rank budget.

\begin{figure}[t]
    \centering
    \includegraphics[width=1\linewidth]{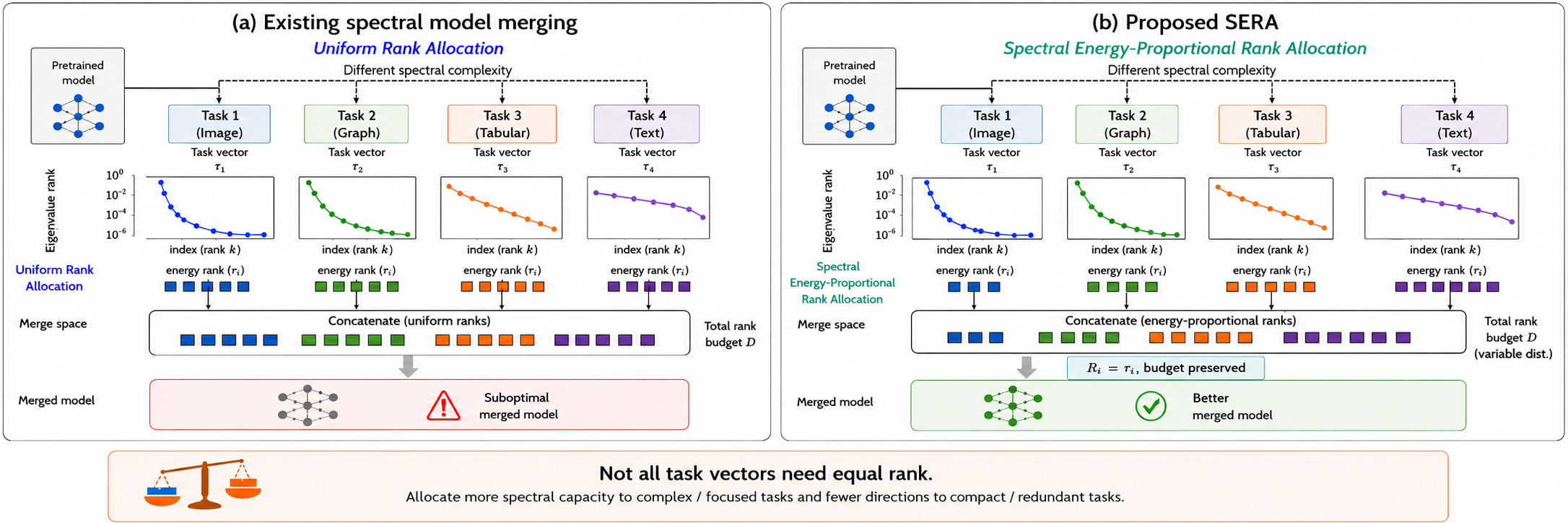}
    \caption{
Overview of our approach. Unlike uniform-rank spectral merging, our method reallocates the same total rank budget according to task-wise spectral energy, giving more capacity to complex tasks and less to compact ones.
}
    \label{fig:scheme}
    \vspace{-0.3cm}
\end{figure}

\section{Method}
\label{sec:method}

We propose \textbf{Spectral Energy-proportional Rank Allocation (SERA)}, a spectral model merging framework that adaptively distributes rank capacity across tasks according to their singular-value energy structure. The key idea is that different task deltas do not require the same number of singular directions to preserve their task-specific information. As shown in Figure~\ref{fig:scheme}, SERA assigns more rank capacity to task deltas whose singular-value energy is distributed over many directions, while assigning fewer ranks to spectrally compact task deltas. This allows the merged model to use a fixed total rank budget more effectively by matching the cover-space capacity to task-wise spectral complexity.

For each task $k$ and layer $l$, let $s_{k,l,j}$ denote the $j$-th singular value of the task delta, and let $q_l$ be the number of available singular directions. We define the raw rank requirement as the minimum number of singular directions needed to preserve a target fraction $t_e \in (0,1)$ of the total spectral energy:
\begin{equation}
r_{k,l}^{\mathrm{raw}}(t_e)
=
\min
\left\{
r :
1 \leq r \leq q_l,\;
\frac{\sum_{j=1}^{r} s_{k,l,j}^{2}}
{\sum_{j=1}^{q_l} s_{k,l,j}^{2}}
\geq t_e
\right\}.
\label{eq:raw_rank}
\end{equation}
This raw rank serves as a task-wise measure of spectral complexity. A task delta with rapidly decaying singular values obtains a small $r_{k,l}^{\mathrm{raw}}$, whereas a task delta with a flatter spectrum obtains a larger $r_{k,l}^{\mathrm{raw}}$.

To isolate the effect of task-adaptive allocation, SERA preserves a fixed total rank budget at each layer. Let $r_{\mathrm{base}}^{(l)}$ denote the reference per-task rank at layer $l$. The total rank budget is defined as
\begin{equation}
B_l = K r_{\mathrm{base}}^{(l)} .
\label{eq:rank_budget}
\end{equation}
SERA keeps this total budget unchanged, but redistributes it across tasks in proportion to their estimated spectral requirements. Therefore, the method changes only how the rank capacity is assigned, not the total amount of spectral capacity used for merging.

\subsection{Spectral Energy-Proportional Rank Allocation}
\label{sec:epr}

\subsubsection{Budget-constrained normalization.}
Given the raw rank requirements, we compute the continuous budget-proportional rank for each task as follows:
\begin{equation}
\widetilde{r}_{k,l}
=
B_l
\frac{r_{k,l}^{\mathrm{raw}}(t_e)}
{\sum_{i=1}^{K} r_{i,l}^{\mathrm{raw}}(t_e)} .
\label{eq:budget_normalization}
\end{equation}
This allocation assigns a larger share of the layer-wise rank budget to tasks with higher spectral complexity, while reducing the allocation for tasks whose energy is concentrated in fewer singular directions.

Since ranks must be integers, we convert the continuous allocation into a feasible integer allocation. We first define the feasible rank set at layer $l$ as
\begin{equation}
\mathcal{A}_l =
\left\{
(r_1,\ldots,r_K)
:
\sum_{k=1}^{K} r_k = B_l,\;
r_k \in \{1,\ldots,q_l\}
\right\}.
\label{eq:feasible_rank_set}
\end{equation}
Then, the final integer allocation is obtained by projecting the continuous ranks onto this feasible set:
\begin{equation}
(r_{1,l}^{\mathrm{alloc}},\ldots,r_{K,l}^{\mathrm{alloc}})
=
\mathop{\mathrm{argmin}}_{(r_1,\ldots,r_K) \in \mathcal{A}_l}
\sum_{k=1}^{K}
\left(
r_k - \widetilde{r}_{k,l}
\right)^2 .
\label{eq:integer_projection}
\end{equation}
This projection preserves the total rank budget while selecting the integer rank vector closest to the continuous energy-proportional allocation. In practice, we implement it by rounding the continuous ranks and then adjusting the remaining surplus or deficit according to the rounding residuals under the budget constraint. The resulting ranks determine how many singular directions each task contributes to the shared spectral representation.

\subsubsection{Adaptive cover space.}

Cover space bases are built using the task-wise ranks allocated by SERA. Instead of selecting the same number of singular directions from every task, SERA selects a different number of directions according to the allocated rank $r_{k,l}^{\mathrm{alloc}}$. Using the notation defined in Sec.~\ref{sec:problem_formulation}, we construct the SERA basis matrices as
\begin{equation}
    \widetilde{U}_{\mathrm{SERA}}^{(l)}
    =
    \left[
    U_{1,l}^{[1:r_{1,l}^{\mathrm{alloc}}]},
    U_{2,l}^{[1:r_{2,l}^{\mathrm{alloc}}]},
    \ldots,
    U_{K,l}^{[1:r_{K,l}^{\mathrm{alloc}}]}
    \right],
    \label{eq:epr_u_basis}
\end{equation}
and
\begin{equation}
    \widetilde{V}_{\mathrm{SERA}}^{(l)}
    =
    \left[
    V_{1,l}^{[1:r_{1,l}^{\mathrm{alloc}}]},
    V_{2,l}^{[1:r_{2,l}^{\mathrm{alloc}}]},
    \ldots,
    V_{K,l}^{[1:r_{K,l}^{\mathrm{alloc}}]}
    \right].
    \label{eq:epr_v_basis}
\end{equation}
Since the allocated ranks satisfy
$\sum_{k=1}^{K} r_{k,l}^{\mathrm{alloc}} = B_l$,
the total number of selected singular directions remains fixed. Thus, spectrally diffuse tasks contribute more directions to the cover space, while spectrally compact tasks contribute fewer directions.

The cover matrices are obtained by taking the orthonormal polar factors of the concatenated bases:
\begin{equation}
    C_U^{(l)}
    =
    \mathrm{polar}
    \left(
    \widetilde{U}_{\mathrm{SERA}}^{(l)}
    \right),
    \qquad
    C_V^{(l)}
    =
    \mathrm{polar}
    \left(
    \widetilde{V}_{\mathrm{SERA}}^{(l)}
    \right).
    \label{eq:epr_cover_matrices}
\end{equation}
The resulting cover space is task-adaptive but budget-preserving. It captures a richer set of directions for tasks with broad spectral energy while avoiding unnecessary capacity consumption by compact tasks.

\subsubsection{Energy Smoothing}
\label{sec:energy_smoothing}

Then, we smooth the retained singular values of each task within its allocated rank block. This reduces the dominance of a few large singular directions while preserving the total retained spectral energy. For task $k$ at layer $l$, we define the shared smoothed singular value as
\begin{equation}
    \bar{s}_{k,l}
    =
    \left(
    \frac{
    \sum_{a=1}^{r_{k,l}^{\mathrm{alloc}}} s_{k,l,a}^{2}
    }{
    r_{k,l}^{\mathrm{alloc}}
    }
    \right)^{1/2}.
    \label{eq:energy_smoothing}
\end{equation}
The smoothed task delta is reconstructed as
\begin{equation}
    \Delta_k^{(l),\mathrm{smooth}}
    =
    U_{k,l}^{[1:r_{k,l}^{\mathrm{alloc}}]}
    \,
    \mathrm{diag}
    \left(
    [
    \bar{s}_{k,l},
    \ldots,
    \bar{s}_{k,l}
    ]
    \right)
    \,
    \left(
    V_{k,l}^{[1:r_{k,l}^{\mathrm{alloc}}]}
    \right)^{\top}.
    \label{eq:smoothed_delta}
\end{equation}

\subsubsection{Projection, aggregation, and reconstruction.}

After constructing the adaptive cover matrices, each task delta is projected into the SERA cover space. Before projection, we use the smoothed task delta $\Delta_k^{(l),\mathrm{smooth}}$. The projected representation of task $k$ is
\begin{equation}
    M_k^{(l)}
    =
    C_U^{(l)\top}
    \Delta_k^{(l),\mathrm{smooth}}
    C_V^{(l)} .
    \label{eq:epr_projection}
\end{equation}
This projection expresses each task update in the same adaptive spectral coordinate system, where the basis dimensions have already been allocated according to task-wise spectral complexity.

We then aggregate the projected task representations using the same aggregation operator. Let
$\mathcal{A}_{\mathrm{TIES}}$ denote the TIES-style sign-consensus aggregation. To preserve the task-wise block structure induced by adaptive rank allocation, we apply a block-diagonal mask:
\begin{equation}
    \overline{M}^{(l)}
    =
    \mathcal{A}_{\mathrm{TIES}}
    \left(
    M_1^{(l)}, \ldots, M_K^{(l)}
    \right)
    \odot
    m^{(l)} .
    \label{eq:epr_aggregation}
\end{equation}
Here, $m^{(l)} \in \{0,1\}^{B_l \times B_l}$ is a block-diagonal mask whose $k$-th diagonal block has size
$r_{k,l}^{\mathrm{alloc}} \times r_{k,l}^{\mathrm{alloc}}$.
More specifically, if
\begin{equation}
    a_{k,l}
    =
    1
    +
    \sum_{i=1}^{k-1}
    r_{i,l}^{\mathrm{alloc}},
    \qquad
    b_{k,l}
    =
    \sum_{i=1}^{k}
    r_{i,l}^{\mathrm{alloc}},
    \label{eq:block_indices}
\end{equation}
then the entries of $m^{(l)}$ are set to one for indices
$(p,q)$ satisfying
$a_{k,l} \leq p,q \leq b_{k,l}$ for some task $k$, and zero otherwise.

Finally, the merged task update at layer $l$ is reconstructed from the adaptive cover space:
\begin{equation}
    \Delta^{\star(l)}
    =
    C_U^{(l)}
    \overline{M}^{(l)}
    C_V^{(l)\top}.
    \label{eq:epr_reconstruction}
\end{equation}
The final merged weight is obtained by adding the scaled merged update to the pretrained weight:
\begin{equation}
    \theta^{\star(l)}
    =
    \theta_0^{(l)}
    +
    \lambda
    \Delta^{\star(l)} ,
    \label{eq:epr_final_weight}
\end{equation}
where $\lambda$ is the global merging scale.

\subsection{Automatic Target Energy Selection}
\label{sec:automatic_te_selection}

The target energy $t_e \in (0,1)$ controls the degree of task-wise rank differentiation in SERA. A smaller $t_e$ focuses on the most dominant singular directions and yields more selective rank allocation, whereas a larger $t_e$ preserves broader spectral information and produces a more balanced allocation. Since the optimal $t_e$ depends on the spectral structure of the fine-tuned models and the number of merged tasks, we select it automatically using a held-out validation split.

We formulate the selection of $t_e$ as a one-dimensional black-box optimization:
\begin{equation}
    t_e^\star
    =
    \arg\max_{t_e \in (0,1)}
    f(t_e),
    \label{eq:te_bo_objective}
\end{equation}
where $f(t_e)$ is the average normalized accuracy of the merged model on the validation split. To reduce the number of candidate evaluations, we use Bayesian Optimization (BO)~\cite{snoek2012practical,jones1998efficient}. BO fits a Gaussian Process surrogate to the observed pairs
$\mathcal{D}=\{(t_e,f(t_e))\}$ and selects the next candidate by maximizing the Expected Improvement acquisition function. The full GP posterior and acquisition formulation are provided in Appendix.

In practice, we warm-start BO with three predefined target energies and stop when the normalized expected improvement becomes smaller than a threshold:
\begin{equation}
    \widetilde{\alpha}_{\mathrm{EI}}
    =
    \frac{
    \max_{t_e} \alpha_{\mathrm{EI}}(t_e)
    }{
    y_{\max} - y_{\min}
    }
    <
    \tau .
    \label{eq:normalized_ei_stop}
\end{equation}
Here, $y_{\max}$ and $y_{\min}$ are the maximum and minimum validation scores observed so far, and we set $\tau=0.01$. The initial points are provided in Appendix.

\vspace{-0.2cm}
\section{Experiments}
\vspace{-0.1cm}

\vspace{-0.1cm}
\subsection{Results of Vision Model Merging}
\vspace{-0.1cm}

We first evaluate the proposed method as a general model merging approach under the standard vision model merging protocol. To ensure a fair and direct comparison with recent SVD-based merging methods, we follow the experimental setting of DC-Merge~\cite{zhang2026dc}, including the benchmark construction, backbone models, fine-tuned checkpoints, evaluation metrics, and baseline methods. The purpose of this experiment is to verify whether replacing uniform rank allocation with our task-adaptive allocation improves merging performance under the same merging budget and evaluation condition.

Following prior works, we conduct experiments in both parameter-efficient and full fine-tuning scenarios. For the parameter-efficient setting, we merge LoRA fine-tuned CLIP \cite{radford2021learning} vision encoders across multiple downstream classification tasks. For the full fine-tuning setting, we merge fully fine-tuned CLIP vision models using the same task collections adopted in DC-Merge. In both cases, we evaluate different CLIP backbones \cite{radford2021learning}, including ViT-B/32 \cite{dosovitskiyimage}, ViT-B/16, and ViT-L/14, to examine whether the proposed rank allocation strategy generalizes across model scales.

We compare our method with representative weight-space and task-vector-based merging baselines, including Weight Averaging \cite{wortsman2022model}, Task Arithmetic \cite{ilharcoediting}, TIES-Merging \cite{yadav2023ties}, Consensus TA \cite{wang2024localizing}, TSV-M \cite{gargiulo2025task}, Iso-CTS \cite{marczak2025no}, and DC-Merge \cite{zhang2026dc}, depending on the corresponding fine-tuning setting. Since our method is designed as an adaptive rank allocation strategy for spectral merging, DC-Merge serves as the most direct baseline. In particular, we keep the total rank budget identical to DC-Merge and only change how the rank capacity is distributed across tasks. This allows us to isolate the effect of task-dependent rank allocation from the effect of using a larger merging subspace.

For evaluation, we report the average accuracy across all merged tasks. We also report normalized accuracy, where each task accuracy is normalized by the performance of its corresponding individually fine-tuned model. This metric reduces the influence of different task difficulty levels and allows a more balanced comparison across heterogeneous datasets. A higher normalized accuracy indicates that the merged model better preserves the task-specific performance of the original fine-tuned models. Overall, this evaluation tests whether our method can improve the average multi-task performance of merged models while maintaining the same experimental protocol and rank budget as existing SVD-based merging methods.

\begin{table*}[t]
\centering
\scriptsize
\setlength{\tabcolsep}{3pt}
\renewcommand{\arraystretch}{1.15}
\resizebox{\textwidth}{!}{%
\begin{tabular}{l|ccc|ccc|ccc}
\hline
\hline
Method
& \multicolumn{3}{c}{ViT-B/32}
& \multicolumn{3}{c}{ViT-B/16}
& \multicolumn{3}{c}{ViT-L/14} \\
\cline{2-4}
\cline{5-7}
\cline{8-10}
& 8 tasks & 12 tasks & 16 tasks
& 8 tasks & 12 tasks & 16 tasks
& 8 tasks & 12 tasks & 16 tasks \\
\hline
Individual
& 87.82 & 88.90 & 87.50
& 89.71 & 90.76 & 89.11
& 92.36 & 93.57 & 92.11 \\

Task Arithmetic \cite{ilharcoediting}
& 52.80 (61.73) & 60.76 (69.12) & 60.04 (68.94)
& 57.70 (65.30) & 64.26 (71.12) & 62.40 (70.16)
& 68.29 (74.38) & 73.69 (78.84) & 69.98 (75.58) \\

KnOTS-TIES \cite{stoica2025model}
& 55.93 (65.03) & 63.03 (71.43) & 61.78 (70.78)
& 60.80 (68.59) & 66.35 (73.38) & 64.31 (72.29)
& 73.61 (79.92) & 75.64 (80.90) & 72.19 (77.98) \\

WUDI-Merging \cite{chengwhoever}
& 55.25 (64.38) & 62.20 (70.64) & 61.24 (70.27)
& 58.95 (66.63) & 65.29 (72.29) & 64.59 (72.51)
& 69.78 (75.91) & 74.25 (79.45) & 71.79 (77.59) \\

TSV-M \cite{gargiulo2025task}
& 58.91 (68.25) & 65.30 (73.86) & 63.51 (72.72)
& 62.97 (70.87) & 68.92 (76.06) & 67.21 (75.35)
& 76.52 (83.00) & 79.67 (85.13) & 74.37 (80.31) \\

Iso-CTS \cite{marczak2025no}
& 63.01 (72.71) & 66.28 (75.01) & 64.61 (74.02)
& 69.06 (77.38) & 71.52 (78.87) & 69.88 (78.22)
& 81.64 (88.31) & 81.35 (86.87) & 77.50 (83.65) \\

DC-Merge \cite{zhang2026dc}
& 64.17 (73.90) & 68.40 (77.22) & 66.27 (75.80)
& 70.53 (78.86) & 73.12 (80.56) & 70.57 (78.91)
& 82.61 (89.42) & 83.62 (89.31) & 79.53 (85.71) \\

\hline
\rowcolor{gray!15}
Ours
& 64.64 (74.30) & 68.59 (77.38) & 66.76 (76.01)
& 70.74 (78.70) & 73.30 (80.70) & 70.95 (79.23)
& 82.95 (89.66) & 84.02 (89.84) & 79.52 (85.88) \\
\hline
\hline
\end{tabular}%
}
\caption{Average absolute accuracy on LoRA-tuned vision model merging. Values in parentheses denote average normalized accuracy. We compare against representative task-vector and subspace-based merging methods under the same evaluation setting.}
\label{tab:lora_vision_merging}
\vspace{-0.7cm}
\end{table*}

Table~\ref{tab:lora_vision_merging} reports the results in the LoRA fine-tuning setting.
Our method achieves the best average accuracy across all CLIP backbones and task scales.
Compared with DC-Merge, our method consistently improves performance under the same total rank budget, showing that the gain comes from task-adaptive rank allocation rather than a larger merging subspace.
The improvements hold for ViT-B/32, ViT-B/16, and ViT-L/14, suggesting that the proposed allocation strategy generalizes across model scales.

\begin{table*}[t]
\centering
\scriptsize
\setlength{\tabcolsep}{3pt}
\renewcommand{\arraystretch}{1.15}
\resizebox{\textwidth}{!}{%
\begin{tabular}{l|ccc|ccc|ccc}
\hline
\hline
Method
& \multicolumn{3}{c|}{ViT-B/32}
& \multicolumn{3}{c|}{ViT-B/16}
& \multicolumn{3}{c}{ViT-L/14} \\
\cline{2-4}
\cline{5-7}
\cline{8-10}
& 8 tasks & 14 tasks & 20 tasks
& 8 tasks & 14 tasks & 20 tasks
& 8 tasks & 14 tasks & 20 tasks \\
\hline
Individual
& 92.83 & 90.88 & 91.37
& 94.64 & 92.76 & 93.17
& 95.81 & 94.29 & 94.73 \\

Weight Averaging \cite{wortsman2022model}
& 66.34 (72.13) & 64.34 (71.12) & 61.04 (67.53)
& 72.22 (76.60) & 69.46 (74.82) & 65.31 (70.36)
& 79.56 (83.15) & 76.73 (81.10) & 71.60 (75.60) \\

Task Arithmetic \cite{ilharcoediting}
& 70.79 (76.55) & 65.32 (72.09) & 60.52 (66.79)
& 75.41 (79.58) & 70.52 (75.89) & 65.78 (70.76)
& 84.93 (88.65) & 79.41 (83.95) & 74.01 (78.07) \\

TIES-Merging \cite{yadav2023ties}
& 75.09 (81.08) & 68.02 (74.83) & 63.38 (69.90)
& 79.74 (84.34) & 73.22 (78.73) & 68.18 (73.26)
& 86.88 (90.69) & 79.46 (84.05) & 75.71 (79.80) \\

Consensus TA \cite{wang2024localizing}
& 75.03 (80.84) & 70.39 (77.36) & 65.43 (71.98)
& 79.39 (83.86) & 74.39 (79.92) & 69.76 (74.93)
& 86.34 (90.08) & 82.22 (86.94) & 79.00 (83.22) \\

TSV-M \cite{gargiulo2025task}
& 85.86 (92.31) & 80.06 (87.88) & 77.07 (84.29)
& 89.01 (93.94) & 84.58 (91.01) & 80.57 (86.45)
& 92.98 (96.98) & 89.17 (91.43) & 87.72 (92.50) \\

Iso-CTS \cite{marczak2025no}
& 86.20 (91.78) & 81.71 (89.70) & 78.05 (85.48)
& 90.91 (95.95) & 86.40 (92.81) & 82.38 (88.36)
& 94.69 (98.81) & 90.98 (96.28) & 90.05 (94.88) \\

DC-Merge \cite{zhang2026dc}
& 87.05 (93.55) & 82.52 (90.62) & 80.58 (88.18)
& 90.78 (95.83) & 87.06 (93.70) & 84.57 (90.76)
& 94.31 (98.38) & 91.01 (96.43) & 90.51 (95.43) \\

\hline
\rowcolor{gray!15}
Ours
& \textbf{87.25} (\textbf{93.92}) & \textbf{83.04} (\textbf{91.34}) & \textbf{80.70} (\textbf{88.36})
& \textbf{91.86} (\textbf{96.27}) & \textbf{87.18} (\textbf{93.90}) & \textbf{84.74} (\textbf{91.02})
& \textbf{94.47} (\textbf{98.56}) & \textbf{91.24} (\textbf{96.70}) & \textbf{90.69} (\textbf{95.63}) \\
\hline
\hline
\end{tabular}%
}
\caption{
Results on vision model merging in the full fine-tuning setting.
We report average accuracy, with average normalized accuracy shown in parentheses.
All baseline results are taken from DC-Merge~\cite{zhang2026dc} under the same evaluation protocol.
}
\label{tab:fft_vision_merging}
\vspace{-0.7cm}
\end{table*}

\begin{table*}[t]
\centering
\scriptsize
\setlength{\tabcolsep}{3pt}
\renewcommand{\arraystretch}{1.15}
\resizebox{\textwidth}{!}{%
\begin{tabular}{c|ccccccccc}
\hline
\hline
\multirow{2}{*}{\textbf{Method}}
& \multicolumn{9}{c}{Datasets} \\
\cline{2-10}
& Cars \cite{krause20133d} & DTD \cite{cimpoi2014describing}  & EuroSAT  \cite{helber2019eurosat}& GTSRB \cite{stallkamp2011german} & MNIST & RESISC45 \cite{cheng2017remote} & SUN397 \cite{xiao2010sun} & SVHN \cite{netzer2011reading} & \textbf{Average} \\
\hline
Task Arithmetic \cite{ilharcoediting}
& 82.0 & 73.6 & 48.8 & 42.1 & 53.1 & 71.5 & 97.5 & 41.2 & 63.7 \\

KnOTS-TIES \cite{stoica2025model}
& 82.7 & 73.7 & 49.3 & 48.9 & 68.9 & 70.9 & 95.5 & 53.8 & 68.0 \\

WUDI-Merging \cite{chengwhoever}
& 82.4 & 73.5 & 48.6 & 46.8 & 54.5 & 72.4 & 96.0 & 46.5 & 65.1 \\

TSV-M \cite{gargiulo2025task}
& 83.9 & 75.1 & 52.4 & 45.5 & 58.3 & 73.1 & 97.6 & 45.3 & 66.4 \\

Iso-CTS \cite{marczak2025no}
& 83.3 & \textbf{84.7} & 49.0 & \textbf{79.6} & 69.4 & \textbf{82.3} & 99.0 & 53.2 & 75.1 \\

DC-Merge \cite{zhang2026dc}
& \textbf{90.9} & 79.6 & 65.5 & 54.0 & 92.6 & 75.0 & 98.2 & 66.8 & 77.8 \\

\hline
\rowcolor{gray!15}
Ours
& 89.9 & 81.5 & \textbf{68.6} & 57.7 & \textbf{94.9} & 76.6 & \textbf{99.5} & \textbf{67.8} & \textbf{79.6} \\
\hline
\hline
\end{tabular}%
}
\caption{Performance on ViT-B/32 8-task benchmark using checkpoints provided by Stoica et al.  \cite{stoica2025model} }
\label{tab:8_task_KnOTS}
\vspace{-0.9cm}
\end{table*}











Table~\ref{tab:fft_vision_merging} reports the results in the full fine-tuning setting.
Our method achieves the best average accuracy across all CLIP backbones and task scales.
The consistent gains show that task-adaptive rank allocation remains effective even for full fine-tuned task vectors, which are typically larger and more prone to inter-task interference.
Because the total rank budget is kept fixed, these results indicate that the distribution of spectral capacity across tasks is a key factor in effective model merging.

A desirable property of a model merging algorithm is robustness to checkpoint-level variation, since independently released fine-tuned models can differ in training recipes, optimization trajectories, and implementation details.
To test whether our method depends on a specific checkpoint source, we conduct additional evaluations using checkpoints released by Stoica et al.~\cite{stoica2025model} and Ilharco et al.~\cite{ilharcoediting}.

\begin{wraptable}{r}{0.6\columnwidth} 
    \vspace{-1.5em} 
    \centering
    \scriptsize
    \setlength{\tabcolsep}{10pt}
    \renewcommand{\arraystretch}{1.15}
    \resizebox{1\linewidth}{!}{
        \begin{tabular}{c|ccc}
        \hline
        \hline
        \textbf{Method} & ViT-B/32 & ViT-B/16 & ViT-L/14 \\
        \hline
        Zeroshot & 48.3 & 55.5 & 64.7 \\
        Individual & 90.5 & 92.6 & 94.2 \\
        \hline
        Task Arithmetic \cite{ilharcoediting} & 70.5 & 74.6 & 84.6 \\
        Fisher Merging \cite{matena2022merging} & 68.3 & 71.7 & 83.7 \\
        RegMean \cite{jin2022dataless} & 71.8 & 76.6 & 82.2 \\
        PCB-Merging \cite{du2024parameter} & 76.3 & 81.5 & 87.5 \\
        TSV-M \cite{gargiulo2025task} & 83.8 & 87.2 & 91.5 \\
        Iso-CTS \cite{marczak2025no} & 84.0 & 88.6 & 92.9 \\
        DC-Merge \cite{zhang2026dc} & 85.1 & 88.8 & 92.7 \\
        \hline
        \rowcolor{gray!15}
        Ours & \textbf{87.1} & \textbf{90.5} & \textbf{94.3} \\
        \hline
        \hline
        \end{tabular}%
    }
    \caption{Performance on eight tasks using checkpoints provided by Ilharco et al. \cite{ilharcoediting}.}
    \label{tab:8_task_task_arithmetic}
    \vspace{-1.5em} 
\end{wraptable}
Table~\ref{tab:8_task_KnOTS} reports results on the ViT-B/32 8-task benchmark using the checkpoints from Stoica et al.~\cite{stoica2025model}.
Our method achieves the highest average normalized accuracy of $79.6\%$, outperforming DC-Merge by $1.8\%p$ and Iso-CTS by $4.5\%p$.
In addition, 
Table~\ref{tab:8_task_task_arithmetic} shows results using checkpoints from Ilharco et al.~\cite{ilharcoediting} across three CLIP backbones.
Our method again achieves the best average accuracy, improving DC-Merge from $85.1$ to $87.1$ on ViT-B/32, from $88.8$ to $90.5$ on ViT-B/16, and from $92.7$ to $94.3$ on ViT-L/14.

These results indicate that the proposed task-adaptive rank allocation strategy is not tied to a specific checkpoint source.
Instead, it remains effective across independently released fine-tuned checkpoints and consistently improves over strong merging baselines under different checkpoint construction pipelines.

\begin{table*}[t]
\centering
\scriptsize
\setlength{\tabcolsep}{3pt}
\renewcommand{\arraystretch}{1.12}
\resizebox{\textwidth}{!}{%
\begin{tabular}{l|cccccccc|c|cccc|c}
\toprule
\multirow{2}{*}{\textbf{Method}}
& \multicolumn{9}{c|}{\textbf{Seen Tasks}}
& \multicolumn{5}{c}{\textbf{Unseen Tasks}} \\
\cmidrule(lr){2-10}
\cmidrule(lr){11-15}
& Cars  \cite{krause20133d} & DTD \cite{cimpoi2014describing} & EuroSAT \cite{helber2019eurosat} & GTSRB \cite{stallkamp2011german} & MNIST \cite{lecun1998gradient} & RESISC45 \cite{cheng2017remote} & SUN397 \cite{xiao2010sun} & SVHN \cite{netzer2011reading} & Average
& CIFAR100 \cite{krizhevsky2009learning}& Flowers \cite{nilsback2008automated} & Pets \cite{parkhi2012cats} & STL10 \cite{coates2011analysis} & Average \\
\midrule
Individual
& 76.61 & 67.34 & 98.19 & 98.29 & 99.15 & 93.97 & 72.49 & 96.50 & 87.82
& -- & -- & -- & -- & -- \\
Zeroshot
& 59.49 & 44.15 & 44.30 & 32.27 & 47.94 & 60.25 & 63.21 & 31.42 & 47.88
& 64.61 & 66.18 & 87.74 & 96.77 & 78.83 \\
\midrule
Task Arithmetic \cite{ilharcoediting}
& 59.82 & 44.31 & 46.52 & 34.71 & 63.43 & 65.78 & 62.61 & 45.25 & 52.80
& 43.83 & 62.30 & 82.12 & 96.14 & 71.10 \\
KnOTS-TIES  \cite{stoica2025model}
& 61.39 & 44.79 & 48.26 & 43.42 & 71.24 & 65.49 & 63.48 & 49.36 & 55.93
& 47.21 & 62.11 & 82.97 & 96.03 & 72.08 \\
WUDI-Merging  \cite{chengwhoever}
& 60.89 & 45.64 & 51.82 & 39.90 & 66.96 & 67.40 & 63.32 & 46.12 & 55.25
& 46.81 & 62.05 & 82.46 & 96.19 & 71.89 \\
TSV-M \cite{gargiulo2025task}
& 62.66 & 46.54 & 55.44 & 46.88 & 75.86 & 69.37 & 63.84 & 50.66 & 58.91
& 51.13 & 61.87 & 84.33 & 96.27 & 73.40 \\
Iso-CTS \cite{marczak2025no}
& 60.72 & \textbf{52.34} & 64.26 & 52.31 & 82.80 & 72.87 & 63.68 & 55.10 & 63.01
& 52.64 & 60.14 & 84.37 & 95.67 & 73.21 \\
DC-Merge \cite{zhang2026dc}
& 62.93 & 50.00 & 61.15 & \textbf{57.27} & \textbf{84.79} & \textbf{75.29} & 64.98 & \textbf{56.93} & 64.17
& 56.90 & 61.44 & 85.08 & 96.00 & 74.86 \\
\hline
\rowcolor{gray!15}
\textbf{Ours}
& \textbf{63.46} & 51.12 & \textbf{64.67} & 57.23 & 83.14 & 75.05 & \textbf{65.87} & 56.87 & \textbf{64.64}
& \textbf{59.06} & \textbf{62.52} & \textbf{85.83} & \textbf{96.34} & \textbf{75.94} \\
\bottomrule
\end{tabular}%
}
\caption{Performance of eight seen tasks and four unseen tasks on ViT-B/32 in LoRA setting. We report absolute accuracy.}
\label{tab:lora_vit_b32_seen_unseen}
\vspace{-0.7cm}
\end{table*}
\begin{table*}[t]
\centering
\scriptsize
\setlength{\tabcolsep}{3pt}
\renewcommand{\arraystretch}{1.12}
\resizebox{\textwidth}{!}{%
\begin{tabular}{l|cccccccc|c|cccc|c}
\toprule
\multirow{2}{*}{\textbf{Method}}
& \multicolumn{9}{c|}{\textbf{Seen Tasks}}
& \multicolumn{5}{c}{\textbf{Unseen Tasks}} \\
\cmidrule(lr){2-10}
\cmidrule(lr){11-15}
& Cars  \cite{krause20133d}& DTD \cite{cimpoi2014describing}  & EuroSAT  \cite{helber2019eurosat} & GTSRB \cite{stallkamp2011german} & MNIST \cite{lecun1998gradient} & RESISC45 \cite{cheng2017remote} & SUN397 \cite{xiao2010sun} & SVHN \cite{netzer2011reading} & Average
& CIFAR100 \cite{krizhevsky2009learning}& Flowers \cite{nilsback2008automated} & Pets \cite{parkhi2012cats} & STL10 \cite{coates2011analysis} & Average \\
\midrule
Individual
& 89.18 & 77.93 & 98.44 & 99.11 & 99.28 & 96.95 & 80.49 & 97.53 & 92.36
& -- & -- & -- & -- & -- \\
Zeroshot
& 77.94 & 55.64 & 64.68 & 50.68 & 76.16 & 71.33 & 68.34 & 58.58 & 65.42
& 76.13 & 79.43 & 93.32 & 99.39 & 87.07 \\
\midrule
Task Arithmetic \cite{ilharcoediting}
& 79.01 & 57.39 & 66.07 & 55.72 & 80.15 & 74.25 & 68.90 & 64.85 & 68.29
& 57.28 & 65.71 & 83.52 & 98.95 & 76.36 \\
KnOTS-TIES  \cite{stoica2025model}
& 81.39 & 60.53 & 71.22 & 65.98 & 89.48 & 79.48 & 69.41 & 71.36 & 73.61
& 57.50 & 72.01 & 90.33 & 98.84 & 79.67 \\
WUDI-Merging  \cite{chengwhoever}
& 79.62 & 58.19 & 66.56 & 57.26 & 85.03 & 76.51 & 68.59 & 66.49 & 69.78
& 60.81 & 69.65 & 87.26 & \textbf{99.03} & 79.19 \\
TSV-M \cite{gargiulo2025task}
& 82.53 & 62.61 & 75.74 & 75.12 & 88.74 & 82.52 & 70.32 & 74.61 & 76.52
& 60.79 & 73.42 & 91.55 & 98.86 & 81.15 \\
Iso-CTS \cite{marczak2025no}
& 81.74 & 67.34 & 84.56 & 88.03 & 96.96 & 86.44 & 69.53 & 78.50 & 81.64
& 61.90 & 72.68 & 91.62 & 98.31 & 81.13 \\
DC-Merge \cite{zhang2026dc}
& \textbf{83.17} & \textbf{68.35} & 84.78 & \textbf{88.48} & 97.06 & \textbf{86.84} & \textbf{71.78} & 80.45 & 82.61
& 65.40 & 72.07 & 91.79 & 98.45 & 81.93 \\
\hline
\rowcolor{gray!15}
\textbf{Ours}
& 82.95 & 68.09 & \textbf{85.93} & 87.16 & \textbf{97.15} & 85.32 & 71.20 & \textbf{81.91} & \textbf{82.95}
& \textbf{67.40} & \textbf{74.61} & \textbf{92.64} & 98.81 & \textbf{83.37} \\
\bottomrule
\end{tabular}%
}
\caption{Performance of eight seen tasks and four unseen tasks on ViT-L/14 in LoRA setting. We report absolute accuracy.}
\label{tab:lora_vit_l14_seen_unseen}
\vspace{-0.7cm}
\end{table*}

\vspace{-0.1cm}
\subsection{Generalization Capability on Unseen Vision Tasks}
\label{sec:exp_lora_generalization}
\vspace{-0.1cm}

We evaluate whether the merged model can generalize beyond the seen tasks.
To validate this, we merge LoRA models trained on eight seen tasks and evaluate the resulting model on four unseen vision tasks: CIFAR100 \cite{krizhevsky2009learning}, Flowers \cite{nilsback2008automated}, Pets \cite{parkhi2012cats}, and STL10 \cite{coates2011analysis}.
The results are reported in Tables~\ref{tab:lora_vit_b32_seen_unseen} and~\ref{tab:lora_vit_l14_seen_unseen}.

Our method consistently improves the average performance on both seen and unseen tasks.
On ViT-B/32, our method improves the seen-task average from $64.17$ to $64.64$ and the unseen-task average from $74.86$ to $75.94$ compared with DC-Merge.
On ViT-L/14, our method further improves the seen-task average from $82.61$ to $82.95$ and the unseen-task average from $81.93$ to $83.37$.
Notably, the gain on unseen tasks is larger than that on seen tasks, suggesting that our method learns a more transferable merged update rather than overfitting to the source tasks.

Across both architectures, our method improves over DC-Merge on all unseen tasks.
These results demonstrate that our method preserves competitive task-merging performance on the seen tasks while providing stronger generalization to unseen visual recognition tasks in the LoRA setting.

\vspace{-0.1cm}
\subsection{Analysis of Task-Adaptive Rank Allocation}
\label{subsec:rank_allocation_analysis}
\vspace{-0.1cm}

\begin{figure*}[t]
\centering
\begin{minipage}{0.48\textwidth}
    \centering
    \includegraphics[width=\linewidth]{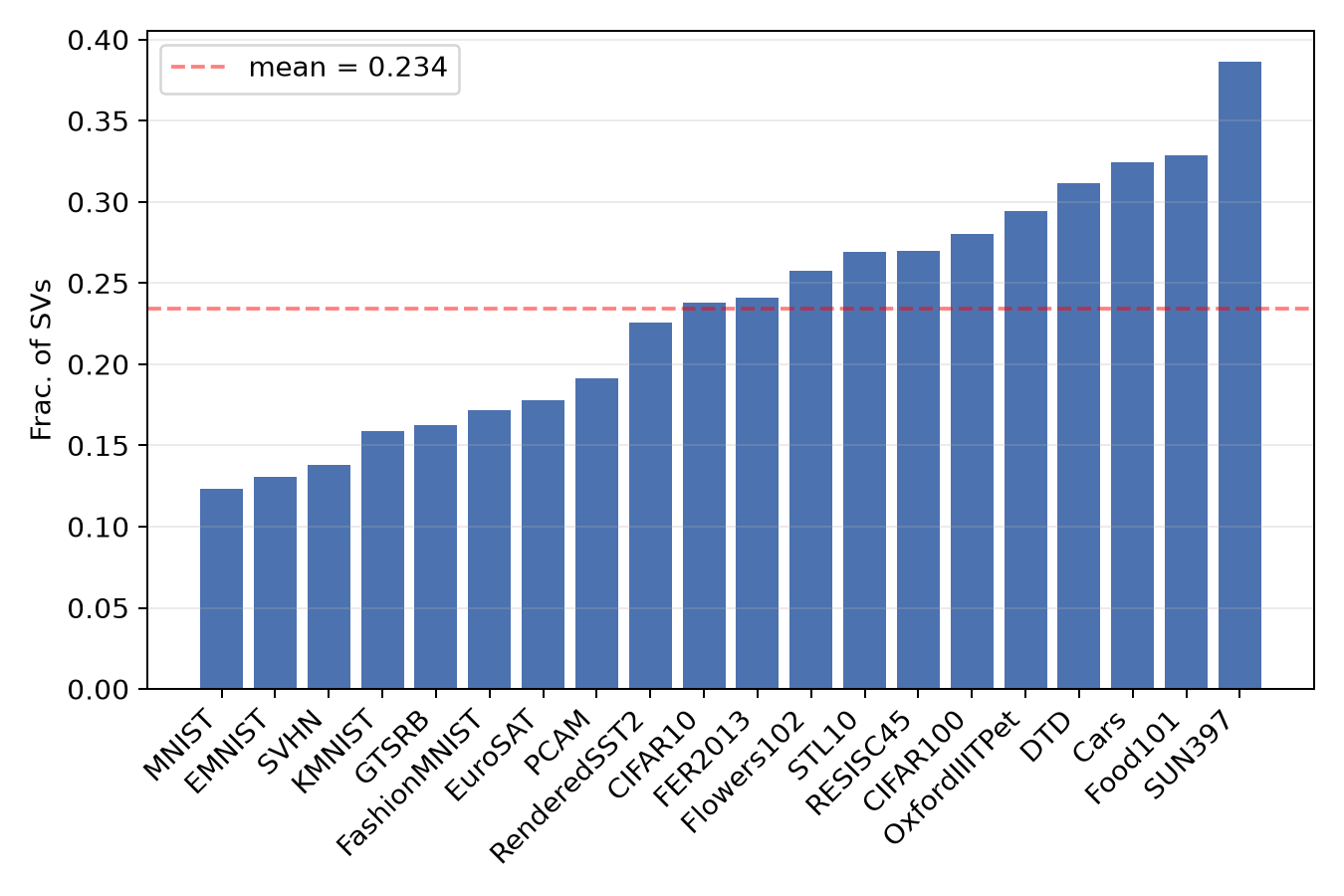}
    \vspace{-0.1cm}
    \centerline{(a) Energy concentration profile}
\end{minipage}
\hfill
\begin{minipage}{0.48\textwidth}
    \centering
    \includegraphics[width=\linewidth]{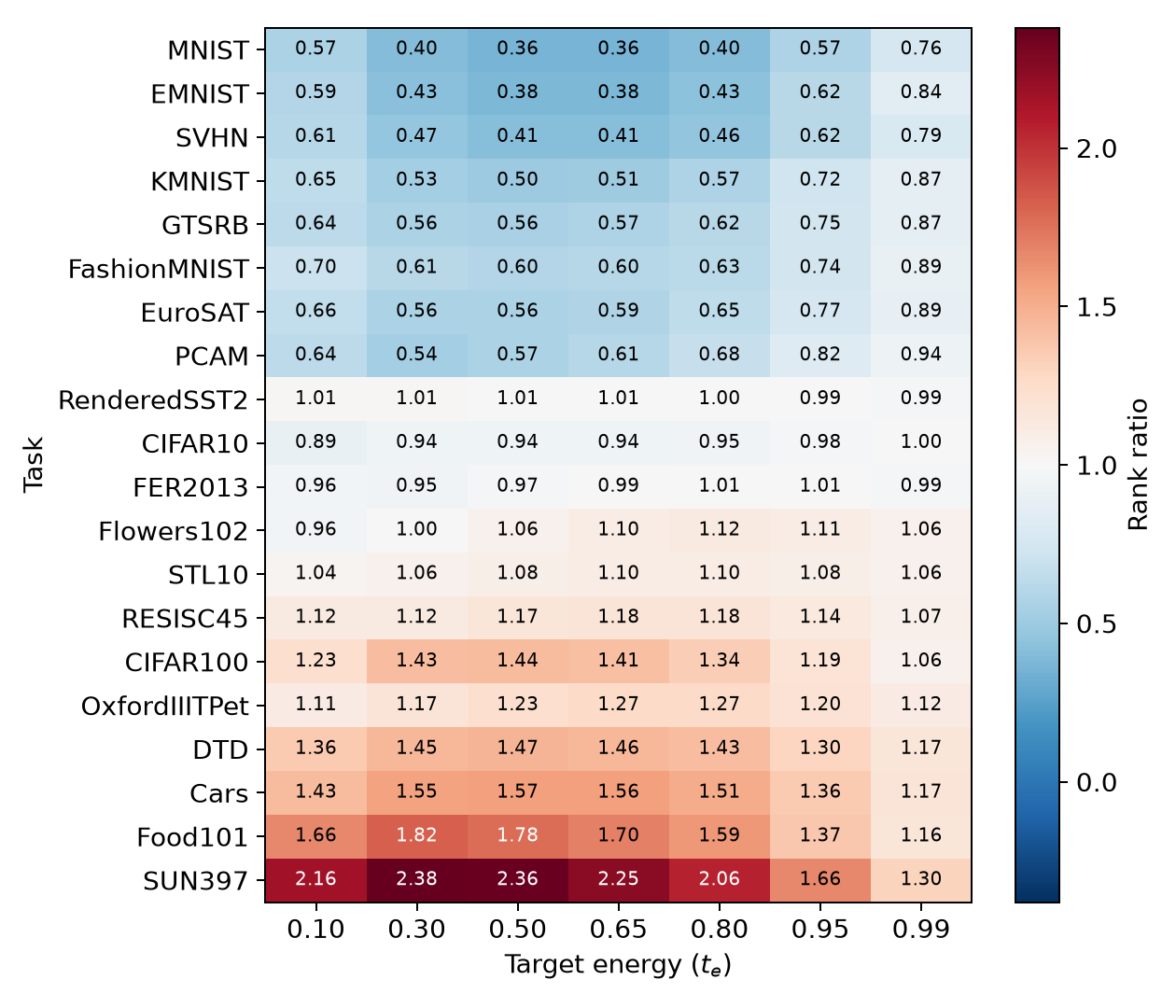}
    \vspace{-0.1cm}
    \centerline{(b) Rank allocation ratio}
\end{minipage}
\caption{
Analysis of task-adaptive rank allocation.
(a) Energy concentration profile across tasks.
(b) Rank allocation ratio relative to uniform allocation.
Blue indicates fewer ranks than uniform allocation, and red indicates more ranks.
}
\label{fig:rank_allocation_analysis}
\vspace{-0.5cm}
\end{figure*}

We analyze how the proposed method distributes the rank budget across tasks.
The motivation for task-adaptive allocation is that different task vectors can have substantially different spectral energy profiles.
If some tasks concentrate most of their energy in a small number of singular directions while others require many singular directions to preserve the same energy level, then uniform rank allocation can be inefficient.

Figure~\ref{fig:rank_allocation_analysis}(a) shows the energy concentration profile of the 20 tasks.
The fraction of singular values required to reach the target energy varies widely across tasks.
For example, digit-related tasks such as MNIST, EMNIST, and SVHN require a relatively small fraction of singular values, indicating highly concentrated spectra.
In contrast, tasks such as Cars, Food101, and SUN397 require a much larger fraction, suggesting more diffuse spectral structure.
This supports the need for task-dependent rank allocation rather than assigning the same rank to every task.

Figure~\ref{fig:rank_allocation_analysis}(b) visualizes the rank allocation ratio of our method relative to uniform allocation.
Blue entries indicate tasks that receive fewer ranks than uniform allocation, while red entries indicate tasks that receive more ranks.
The allocation pattern aligns with the energy concentration profile: tasks with compact spectra receive fewer ranks, whereas tasks with diffuse spectra receive more ranks.
For instance, MNIST, EMNIST, and SVHN are consistently assigned fewer ranks, while SUN397, Food101, Cars, and DTD receive more rank capacity.
This provides empirical evidence that the proposed method performs spectral-energy-aware allocation rather than applying a fixed rank budget uniformly across tasks.

Overall, this analysis explains why task-adaptive rank allocation improves merging performance.
By reallocating rank capacity from spectrally compact tasks to spectrally diffuse tasks, our method uses the same total rank budget more effectively and constructs a more balanced merging subspace.

\vspace{-0.1cm}
\subsection{Analysis of Target Energy Sensitivity}
\label{subsec:target_energy_sensitivity}
\vspace{-0.1cm}

\subsubsection{Relation between the optimal $t_e^\ast$ and Task Count}
\label{subsubsec:target_energy_shift}

\begin{figure}[htbp]
    \begin{minipage}[t]{0.48\columnwidth}
        \centering
        \includegraphics[width=1.\linewidth]{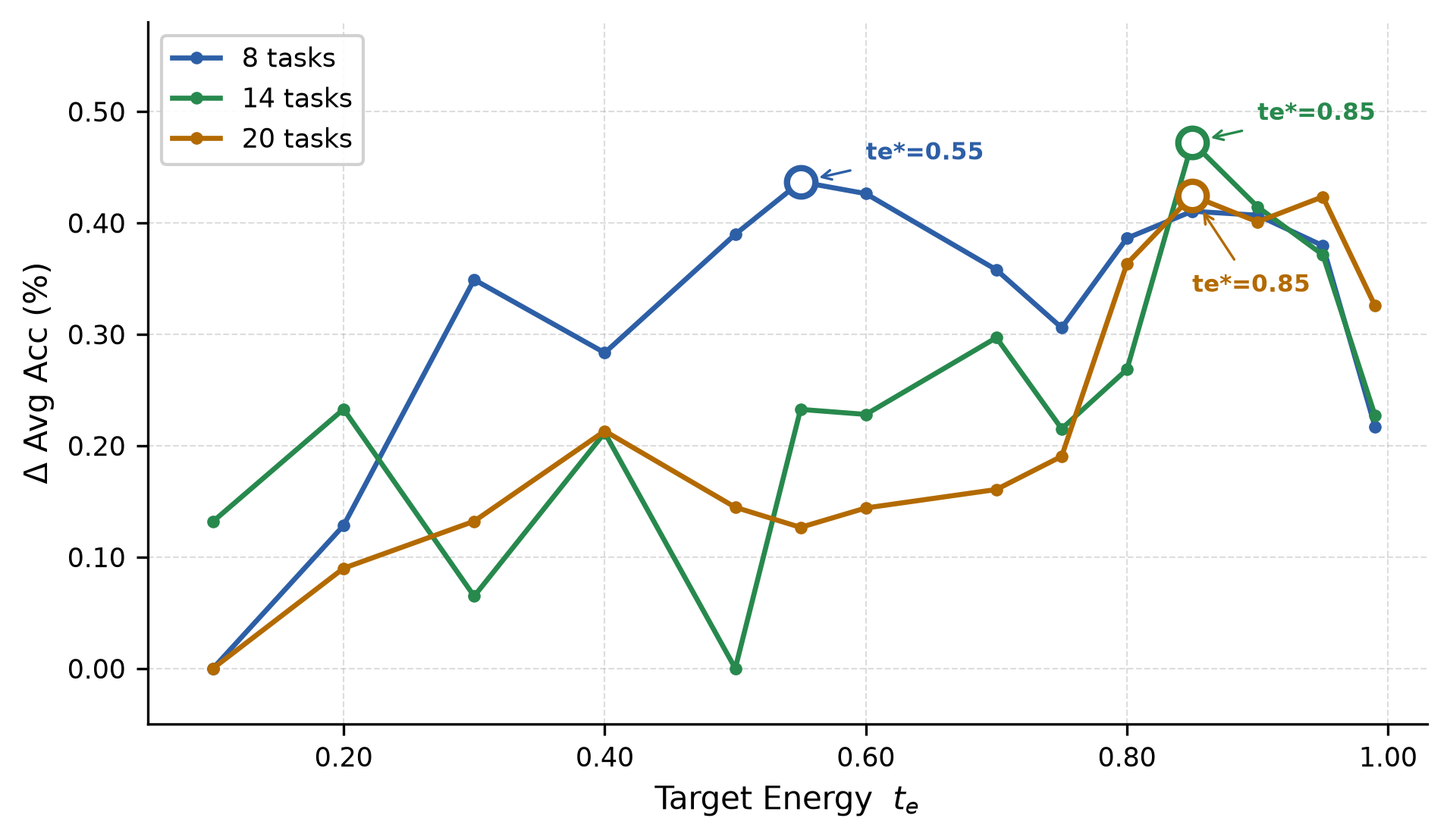}
        \caption{Average accuracy gain over the DC-Merge baseline versus target energy $t_e$ on ViT-B/32 with 8, 14, and 20 tasks.}
        \label{fig:target_energy_sensitivity}
    \end{minipage}
    \hfill
    \begin{minipage}[t]{0.48\columnwidth}
        \centering
        \includegraphics[width=1.\linewidth]{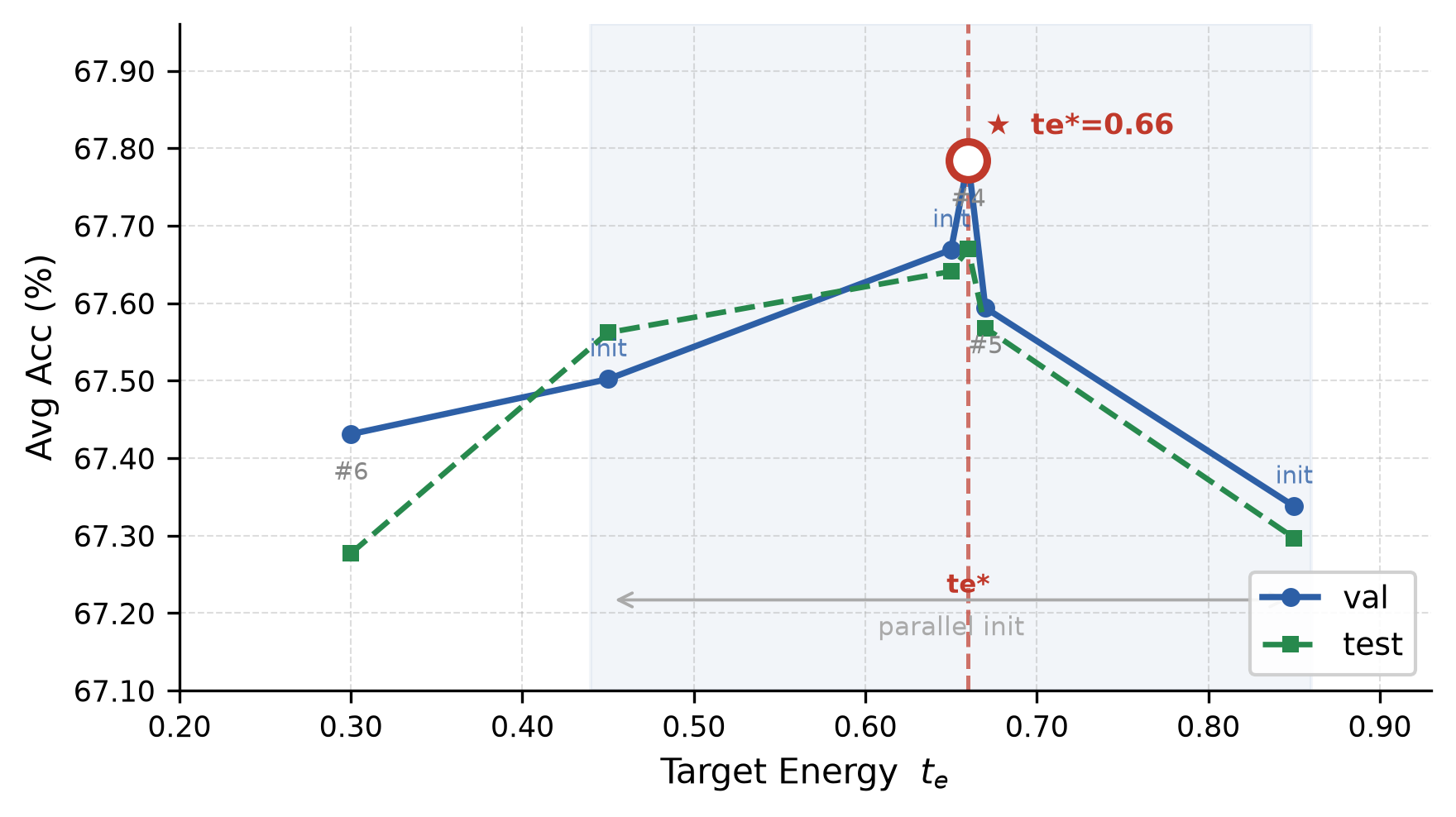}
        \caption{Ablation about $t_e$ on ViT-B/32 with 8 tasks.}
        \label{fig:bo_convergence}
    \end{minipage}
    \vspace{-0.9cm}
\end{figure}

We evaluate SERA across target energies $t_e \in [0.10, 0.99]$ on ViT-B/32 FFT
checkpoints merged from DC-Merge's cover space, varying the task count to 8, 14, and 20.
Figure~\ref{fig:target_energy_sensitivity} reports the accuracy gain over the DC-Merge
baseline at each $t_e$.

Two findings stand out.
First, SERA consistently outperforms DC-Merge across the full range, confirming that
energy-proportional rank allocation is beneficial regardless of the specific $t_e$ chosen.
Second, the optimal $\tilde{t}_e^{*}$ exhibits a clear rightward shift as task count grows:
$\tilde{t}_e^{*} \approx 0.55$ for 8 tasks, rising to $\tilde{t}_e^{*} \approx 0.85$ for 14 and 20 tasks.
This shift reflects the growing inter-task interference at higher task counts: retaining a larger
fraction of singular-value energy per task is necessary to preserve task-specific directional
structure within the shared cover space.


\subsubsection{Bayesian Optimization Converges in Few Evaluations}
\label{sec:bo_convergence}

The task-count dependence of $\tilde{t}_e^{*}$ precludes the use of a single fixed default.
Figure~\ref{fig:bo_convergence} traces the BO trajectory for ViT-B/32 LoRA checkpoints
(8 tasks merged, evaluated on 12 tasks: 8 seen + 4 unseen).


Starting from three parallel evaluations at $t_e \in \{0.45, 0.65, 0.85\}$, the GP surrogate identifies
$\tilde{t}_e^{*} = 0.66$ in only one additional step.
Two further explorations confirm convergence, reaching the stopping criterion
$d_{\mathrm{EI}} < \tau$ at the sixth evaluation.
Crucially, the validation-optimal $\tilde{t}_e^{*}$ coincides exactly with the test-optimal point:
both val and test curves peak at $t_e = 0.66$, confirming that the validation proxy generalises
reliably to unseen tasks without overfitting.


\vspace{-0.1cm}
\section{Conclusion}
\vspace{-0.1cm}
In this paper, we proposed a task-dependent rank allocation strategy for spectral model merging. Unlike conventional weighted averaging or uniform-rank merging, our method assigns rank capacity according to the spectral structure of each task vector. This allows complex or isolated tasks to use more singular directions, while compact or redundant tasks are represented with fewer directions. By incorporating task-wise spectral complexity, the proposed method preserves the benefits of low-rank merging while using the shared rank budget more effectively. Our experiments show that adaptive rank allocation consistently improves multi-task merging performance over uniform allocation, suggesting that effective model merging should consider not only how task vectors are aggregated, but also how much spectral capacity each task requires.


%
%
\bibliographystyle{splncs04}
\bibliography{main}
\end{document}